# Visual object categorization with new keypoint-based adaBoost features

Taoufik Bdiri, Fabien Moutarde, and Bruno Steux

*Abstract*— We present promising results for visual object categorization, obtained with adaBoost using new original "keypoints-based features". These weak-classifiers produce a boolean response based on presence or absence in the tested image of a "keypoint" (a kind of SURF interest point) with a descriptor sufficiently similar (i.e. within a given distance) to a reference descriptor characterizing the feature. A first experiment was conducted on a public image dataset containing lateral-viewed cars, yielding 95% recall with 95% precision on test set. Preliminary tests on a small subset of a pedestrians database also gives promising 97% recall with 92 % precision, which shows the generality of our new family of features. Moreover, analysis of the positions of adaBoost-selected keypoints show that they correspond to a specific part of the object category (such as "wheel" or "side skirt" in the case of lateral-cars) and thus have a "semantic" meaning. We also made a first test on video for detecting vehicles from adaBoost-selected keypoints filtered in real-time from all detected keypoints.

## I. INTRODUCTION AND RELATED WORK

One of the key features for enhancing safety in intelligent vehicles is efficient and reliable detection of surrounding moving objects such as pedestrians and vehicles. It is particularly interesting to be able to properly detect laterally incoming cars that could lead to lateral collisions.

Many techniques have been proposed for visual object detection and classification (see e.g. [3] for a review of some of the state-of-the-art methods for pedestrian detection, which is the most challenging). Of the various machine-learning approaches applied to this problem, only few are able to process videos in real-time. Among those, the boosting algorithm with feature selection was successfully extended to machine-vision by Viola & Jones [2].

The adaBoost algorithm was introduced in 1995 by Y. Freund and R. Shapire [1], and its principle is to build a *strong classifier*, assembling weighted weak classifiers, those being obtained iteratively by using successive weighting of the examples in the training set. Most published works using adaBoost for visual object class detection are using the Haar-like features initially proposed by Viola & Jones for face and pedestrian detection.

However, adaBoost outcome may strongly depend on the family of features from which the weak classifiers are drawn. Recently, several teams [4][5] have reported interesting results with boosting using other kinds of features directly inspired from the Histogram of Oriented Gradient (HOG) approach. Our lab has been successfully investigating boosting with pixel-comparison-based features named "control-points" (see [6] for original proposal, and [7] for recent results with a new variant).

In the present work we investigate boosting of "keypoint presence features", where "keypoint" are a variant of SURF points implemented in our lab (see below), and already successfully applied to real-time person re-identification [11]. To our knowledge, the idea of using interest point descriptors as boosting features was first proposed by Opelt et al. in [8], but it was in a more general framework, and they were considering SIFT points and descriptors [9] which are quite slow to compute, compared to the SURF points and descriptors [10].

The paper is organized as follows: in section II we briefly present the principle of the "Camellia keypoints" we are using; section III explains how we use keypoints to define a new original family of weak classifiers, and how is realized the feature-selection in this family during each boosting step; section IV presents experimental results on a publicly available image dataset of laterally-viewed cars, and preliminary evaluation on a small pedestrians dataset; section V presents our first step in building an original object detection scheme that could be used with our particular keypoint-based classifier; and section VI draws some conclusions and perspectives.

## II. CAMELLIA "KEYPOINTS"

The interest point detection and descriptor computation is performed using "key-points" functions available in the Camellia (http://camellia.sourceforge.net) image processing library. These Camellia key-points detection and descriptor functions – named CamKeypoints - implement a variant of SURF [10]. SURF itself is an extremely efficient method (thanks to the use of integral images) inspired from the more classic and widely used interest point detector and descriptor SIFT [9].

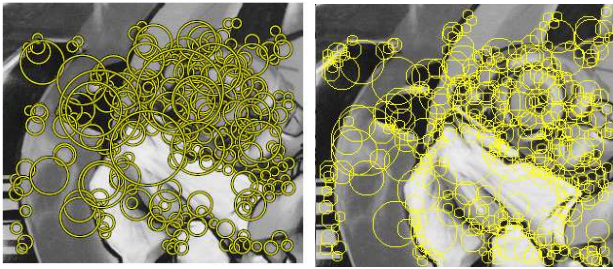

*Fig. 1. SURF interest points (left) v.s. Camellia keypoints (right); they are very similar except for voluntary suppression of multiple imbricated blobs at different scales (cf. upper left).*

As for SURF interest points, the detection of Camellia keypoints is a "blob detector" based on finding local Hessian maxima, those being efficiently obtained by approximating second order derivatives with box filters computed with integral image. Our keypoints are however not exactly the same as SURF points (as can be seen on figure 1), in particular because multiple imbricated blobs at various scales are voluntarily avoided. In contrary to SURF and SIFT, CamKeypoint scale selection is not based on overlapping octaves, but on a set of discrete scales from which the scale of a keypoint is derived by quadratic interpolation. This speeds up the keypoints detection wrt. to SURF by a factor of 2 without sacrificing the quality of scale information, as was shown by some experiments.

The descriptor used for each Camellia keypoint is similar to the SURF descriptor : an image patch corresponding to the keypoint location and scale is divided in 4x4=16 sub-regions, on each of which are efficiently computed (by using integral image approach) the following 4 quantities:

$$\sum dx$$
$$\sum |dx|$$
$$\sum dy$$
$$\sum |dy|$$

where *dx* and *dy* are respectively the horizontal and vertical gradient. The total descriptor size is therefore 16 x 4 = 64. In order to avoid all boundary effects in which the descriptor abruptly changes when a keypoint physically changes, bi-linear extrapolation is used to distribute each of the 4 quantities above into 4 sub-regions. Experiments have shown that this really improves the quality of the descriptor wrt. SURF. In addition to this, CamKeypoints support color images by adding 32 elements of gradient information by color channel (U and V) to the signature, resulting in a 128 descriptor size for color descriptors.

Another main difference between Camellia Keypoints and SURF lies in that the Camellia implementation uses integer-only computations – even for the scale interpolation –, which makes it even faster than SURF, and particularly well-suited for potential embedding in camera hardware. SIFT and SURF make extensive use of floating point computations, which makes these algorithms power hungry.

## III. ADABOOST WITH "KEYPOINT PRESENCE" FEATURES

Our object recognition approach uses the same general feature-selecting boosting framework as pioneered by Viola&Jones in [2]. The originality of our work is to define and use as weak classifiers a new original feature family, instead of Haar features. This new feature type is a weak classifier that answers positively on an image if and only if, among all the Camellia keypoints detected in the image, there is at least one of them whose descriptor is similar enough to the "reference keypoint descriptor" associated with the weak-classifier.

More formally, each "keypoint presence" weak-classifier is defined by a keypoint SURF descriptor D in $\Re^{64}$, and a descriptor difference threshold scalar value d. This weak-classifier h(D,d,I) answers positively on an image I if and only if I contains at least one keypoint whose descriptor D' is such that $|D-D'|<d$, where the "sum of absolute difference" (SAD) L1-distance is used: if two keypoints $K_1$ and $K_2$ respectively have $\{Desc_1[i], i = 1…64\}$ and $\{Desc_2[i], i = 1…64\}$, then, the distance between K1 and K2 is given by equation 1 below:

$$Dist(K_1, K_2) = \sum_{i=1}^{64} abs(Desc_1[i] - Desc_2[i]) \quad (1)$$

The rationale of boosting "keypoint presence features" for image categorization is that it should be possible, for a given object category, to determine a set of characteristic interest points whose simultaneous presence would be representative of that particular category. This is similar in spirit, but with a completely different algorithm, to the "part based" approach proposed by [12].

The training method is the standard feature-selecting adaBoost algorithm, in which, at each boosting step, the SURF descriptor D is chosen among all descriptors found in *positive* example images. More formally, let the training set be composed of positive images Ip1, Ip2, …., and of negative images In1, In2, …We first apply the Camellia keypoint detector on all *positive* images Ip1, Ip2, …, and build the "positive keypoints set" Spk = $\{K_1, K_2, K_3, …, K_Q\}$ as the union of all Camellia keypoints detected on any positive examples of the training set. The adaBoost feature-selection has to select, at each boosting step, a particular "keypoint presence" weak classifier defined by a 64D descriptor and a scalar threshold. The descriptor will be chosen among those of positive keypoints collected in Spk.

In order to choose a threshold value, we apply keypoints detection on all negative images as well, so that we can compare descriptors of the positive keypoints in Spk to descriptors of all keypoints found in training images. We define the "distance" between any given keypoint K and any given image I as the smallest descriptor difference between K and all keypoints KIj found in image I:

$$dist(K,I) = \min_{KIj \text{ keypoint found in image I}} \{ dist(K, KI_j) \} \quad (2)$$

where dist(K,KIj) is the SAD of descriptors as defined in equation (1). This allows us to build a matrix M of distances between positive keypoints and all training images, where Mij = dist(Ki , Ij). As illustrated on figure 2, this QxN matrix (with Q the number of positive keypoints and N the number of training images) has at least one zero on each line, on the column corresponding to the positive image in which the keypoints was found.

$$\begin{pmatrix} I_{p1} & I_{p2} & I_{p3} & I_{n1} & I_{n2} & I_{n3} \\ 0 & x & x & x & x & x \\ 0 & x & x & x & x & x \\ x & 0 & x & x & x & x \\ x & 0 & x & x & x & x \\ x & 0 & x & x & x & x \\ x & x & 0 & x & x & x \\ x & x & 0 & x & x & x \\ x & x & 0 & x & x & x \\ x & x & 0 & x & x & x \end{pmatrix}$$

*Fig. 2. Matrix of distances between keypoints found on positive image example (one for each row) and all N training images (positives and negatives, one for each column)*

We make a growing sorting of the distance matrix M, row by row, and then we take the *middle of each two successive distances* in the sorted matrix to build the set {$T_{ik}$, k=1,…,N} of candidate threshold values for a feature testing presence of the corresponding positive keypoint $K_i$.

At each boosting step, we choose among all (Ki,Tik) couples the one that gives the lowest weighted error on the training set: (i*,k*) = argmin$_{ik}$ ( $\sum_{j=1}^{N} w_j |h(K_i,T_{ik},I_j) - l_j|$ ), and the selected weak classifier is h($K_{i*},T_{i*k*}$, . ).

## IV. EXPERIMENTS AND RESULTS

### A. Lateral cars database

For a first evaluation of our approach, we used the publicly available (http://l2r.cs.uiuc.edu/~cogcomp/Data/Car/) lateral-car dataset collected by Agarwal et al. [12]. This database contains 550 positive images and 500 negative images. For training, we use 352 positive images, and 322 negative images, the rest being used as a test set for evaluation. Note that the partition between training and testing subset is random. Some examples from the training set are shown on figure 3.

Figure 4 shows the typical error evolution during adaBoost training: as is usual with boosting, the training error quickly falls to zero, and the error on test set continues to diminish afterwards. This shows that boosting by assembling features extracted from our new "keypoint presence" family does work and allow to build a strong classifier able to discriminate a given object category. On this particular case, there seems to be no clear improvement on test dataset for boosting steps T>150.

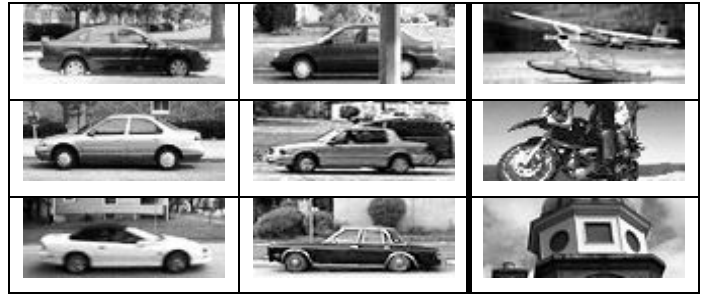

*Fig.3. Some positive (2 left columns) end negative (right column) examples from the training set*

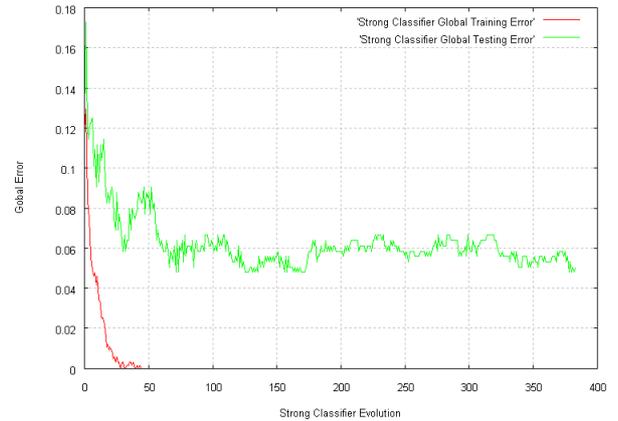

*Fig.4. Typical evolution, during successive boosting steps, of errors on training and test*

Figure 5 shows the precision-recall curve, computed on the independent test set, for boosted strong classifiers with respectively 10 and 300 "keypoint presence" weak-classifiers assembled. The classification result is very good, with a recall of ~95% for a precision of ~95%.

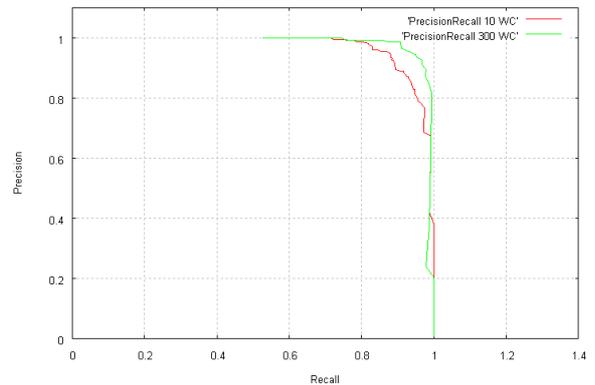

*Fig. 5. Precision-recall curve computed on test set, for strong boosted classifier assembling 10 and 300 weak classifiers selected from our new "keypoint presence" family.*

In order to further analyze how the obtained classifier works, we looked at the evolution of strong classifier output on test images as a function of the boosting step. As can be seen on figure 6a, we typically obtain, on positive images, only positive votes by the first few weak-classifiers, and then some negative votes decrease the global output, but the weighted vote remains largely above the 0.5 threshold for positive classification.

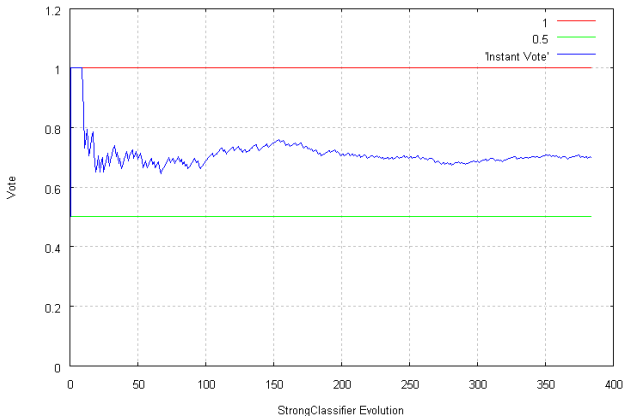

*Fig. 6a. Evolution with increasing boosting steps of strong classifier output on a given positive image*

The typical strong classifier output evolution on negative image is roughly symmetric, as illustrated on figure 6b.

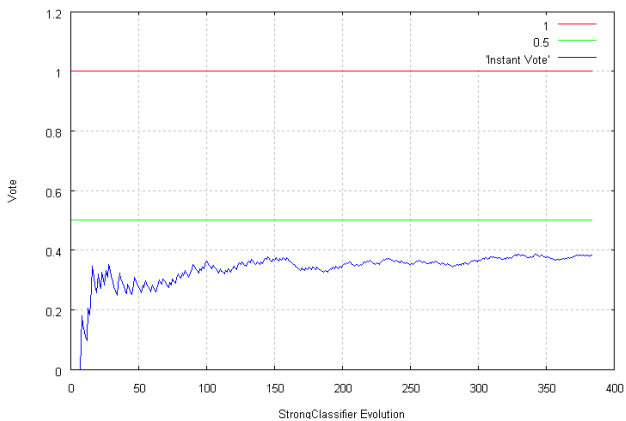

*Fig. 6b. Evolution with increasing boosting steps of strong classifier output on a given negative image*

We also checked how the boosting-selected "keypoint presence" features respond on positive and negative images. As illustrated on figure 7, some of the adaboost-selected features vote positive on negative images, but this does not prevent correct classification as negative by the strong classifier.

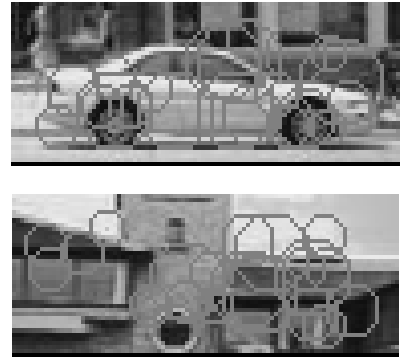

*Fig. 7. Illustration on one positive image and one negative image of the positively responding adaboost-selected keypoints; some of them do vote positive on some negative images, but the strong classifier still correctly classifies those negatives images.*

### B. Pedestrians database

As a quick check for the generality of our new family of features, we have applied our method to a *small subset* of the publicly available pedestrians database collected by Munder and Gavrila [3]. For training computation time reasons, we used only 550 positive images and 500 negative images from their first training set, and split them as 2/3 for our training, and 1/3 for our testing.

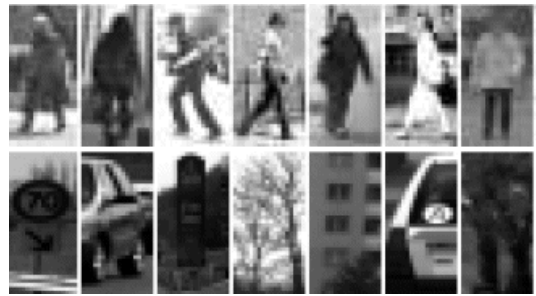

*Fig. 8. Some examples from the pedestrians database subset.*

The boosting with our new family of features indeed learns normally, as can be checked on the evolution with boosting steps of training and testing errors shown on figure 9.

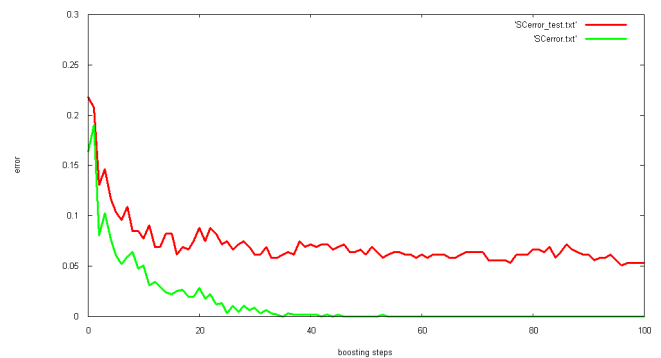

*Fig. 9. Evolution during successive boosting steps, of the training and testing errors on the subset of pedestrians database.*

The precision-recall curve computed on the test set is also correctly evolving to the upper-right corner, attaining a good 97% recall / 92% precision with a 100-features strong classifier, as can be seen on figure 10. However, the classification performance of our method on the much bigger full-sized pedestrians database still remains to be evaluated.

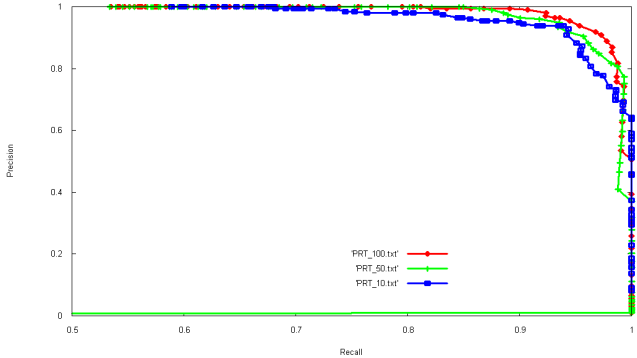

*Fig. 10. Precision-recall curves computed on test set, at boosting steps 10 (lower curve), 50 (middle curve) and 100 (upper curve).*

Finally, we illustrate on figure 11 what are the adaboost-selected keypoints replying "positive" on some typical positive examples. It can be noticed that some keypoints typically seem to circle the head, others the shoulder, and others the "upper inter-leg" part

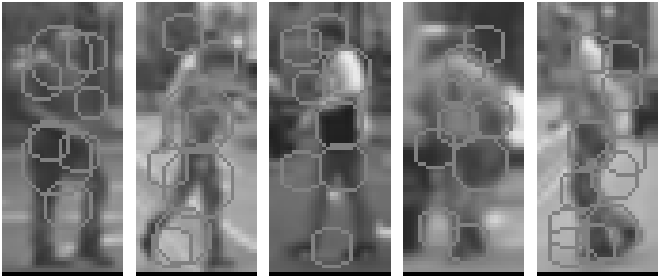

*Fig.11. Illustration on some positive examples of the positively responding adaboost-selected keypoints*

V. OBJECT DETECTION FROM KEYPOINTS

There are several motivations for our new feature type. One is that a classifier based on the simultaneous presence of several characteristic keypoints matches the intuition we can have on how human do categorize image by spotting some characteristic parts. In order to check if our adaBoost-selected keypoints make sense from this point of view, we decided to check on positive images where are located the "positively responding keypoints" for a given feature of the strong classifier.

Figure 12 illustrates the positions of all keypoints, cumulated on all positive example images, that are within the descriptor distance threshold of one given adaBoost-selected keypoints.

This clearly shows that the keypoints selected correspond to specific parts of the object category, such as the wheels or the side skirt, which means they have a semantic signification relative to the object category.

| Weak Classifier | Image 1 | Image 2 | Specificity |
|---|---|---|---|
| 40 | | | Keypoints concentrated on the rectangular side of the car |
| 230 | | | Keypoints concentrated under the car between the two wheels |
| 300 | | | Keypoints concentrated on wheels |

*Fig. 12. Position of adaBoost-selected positively responding keypoints cumulated on all positive example images.*

Another motivation for these new kind of adaBoost features is that, by nature of the features, it should be possible to derive the localizations in the image of objects of the searched category quite straightforwardly by some kind of clustering, or possibly a Hough-like method, applied to the positions of positively-responding keypoints, thus making costly window-scanning unnecessary.

As a first test, we computed all keypoints on a video, and filtered them to keep only the positively-responding ones, as illustrated on figure 13, where one can see that laterally incoming car on upper-right part of field is rather well delineated as a single group of positive keypoints.

Note that the computation of all keypoints, as well as their filtering for keeping only the positively-responding ones is done *in real-time* on the video.

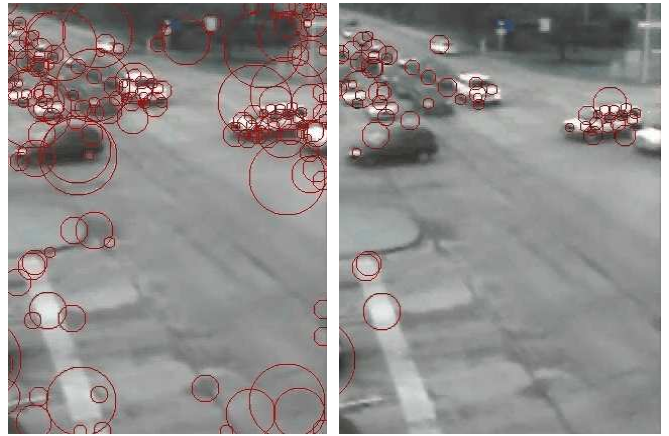

*Fig. 13. First detection test on a video: all keypoints on the left side, and only positively-reponding keypoints on the right side.*

## VI. CONCLUSIONS AND PERSPECTIVES

We have presented a new family of weak-classifiers, "keypoint presence features", to be used for boosting for object category visual recognition. We have obtained first successful test of boosting "keypoint presence features", applied to lateral car recognition, yielding 95% recall with 95% precision on test set. Moreover, analysis of the positions of adaBoost-selected keypoints show that they correspond to a specific part of the object category (such as "wheel" or "side skirt") and thus have a "semantic" meaning. Preliminary test on a small subset of a pedestrians database also gives promising results, showing that our new family of features can be used for recognition of various types of object categories.

Perspectives include tests on other datasets, in particular for other object categories. Also, an optimization of the keypoint-threshold selection is underway, as the current version makes training rather computer-intensive for large datasets.

More importantly, we are currently developing an object-localization method based on the analysis of positions of positively-responding keypoints. Finally, we are considering exploiting the relative positions of keypoints, instead of only their simultaneous presence, for further improvement of the performances.